\documentclass[11pt,a4paper]{article}
\usepackage[hyperref]{conll-2019}
\usepackage{times}
\usepackage{todonotes}
\usepackage{booktabs}
\usepackage{xcolor}
\usepackage{graphicx}
\usepackage{latexsym}
\usepackage[ruled,vlined]{algorithm2e}
\usepackage[T1]{fontenc}
\usepackage{subcaption}
\usepackage{algpseudocode}

\usepackage{url}

\aclfinalcopy %

\title{On Model Stability as a Function of Random Seed}

\author{Pranava Madhyastha \\
  Department of Computing\\
  Imperial College London\\
  {\tt pranava@imperial.ac.uk} \\\And
  Rishabh Jain\thanks{~~This work was conducted when the author was a student at Imperial College London.}\\
  Bloomberg \\
  London\\
  {\tt rjain213@bloomberg.net} \\}

\date{}

\begin{document}
\maketitle
\begin{abstract}
In this paper, we focus on quantifying model stability as a function of random seed by investigating the effects of the induced randomness on model performance and the robustness of the model in general. We specifically perform a controlled study on the effect of random seeds on the behaviour of attention, gradient-based and surrogate model based (LIME) interpretations. Our analysis suggests that random seeds can adversely affect the consistency of models resulting in counterfactual interpretations. We propose a technique called~\emph{Aggressive Stochastic Weight Averaging (ASWA)} and an extension called~\emph{Norm-filtered Aggressive Stochastic Weight Averaging (NASWA)} which improves the stability of models over random seeds. With our ASWA and NASWA based optimization, we are able to improve the robustness of the original model, on  average reducing the standard deviation of the model's performance by $72\%$. 
\end{abstract}

\section{Introduction}

There has been a tremendous growth in deep neural network based models that achieve state-of-the-art performance. In fact, most recent end-to-end deep learning models have surpassed the performance of careful human feature-engineering based models in a variety of NLP tasks. However, deep neural network based models are often brittle to various sources of randomness in the training of the models. This could be attributed to several sources including, but not limited to, random parameter initialization, random sampling of examples during training and random dropping of neurons. It has been observed that these models have, more often, a set of \emph{random seeds} that yield better results than others. This has also lead to research suggesting random seeds as an additional hyperparameter for tuning \cite{bengio2012practical}\footnote{\url{http://www.argmin.net/2018/02/26/nominal/}}.
One possible explanation for this behavior could be the existence of multiple local minima in the loss surface.  This is especially problematic as the loss surfaces are generally non-convex and may have multiple saddle points making it difficult to achieve model stability.%
\begin{figure}[h]
    \centering
    \resizebox{0.9\linewidth}{!}{
    \begin{tabular}{p{\linewidth}}\toprule
         if high crimes were any more generic it would have a universal \colorbox{pink}{product} code \colorbox{red}{instead} of a title \begin{center}($\Pr{(Y_{{negative}})}=0.99$)\end{center}\\\midrule
         if \colorbox{magenta}{high} \colorbox{red}{crimes} \colorbox{pink}{were} any more generic it would have a universal product code instead of a title \begin{center}($\Pr{(Y_{{negative}})=0.98}$)\end{center}\\\bottomrule
    \end{tabular}
    }
    \caption{Importance based on attention probabilities for two runs of the same model with \textbf{same parameters and same hyperparameters}, but with \textbf{two different random seeds} (color magnitudes: pink$<$magenta$<$red)}
    \label{fig:introeg}
\end{figure}

Recently the NLP community has witnessed a resurgence in interpreting and explaining deep neural network based models~\cite{jain2019analysis,DBLP:journals/corr/abs-1902-10186, alvarez2017causal}. Most of the interpretation based methods involve one of the following ways of interpreting models: a) sample oriented interpretations: where the interpretation is based on changes in the prediction score with either upweighting or perturbing samples~\cite{jain2019analysis,DBLP:journals/corr/abs-1902-10186,koh2017understanding}; b) interpretations based on feature attributions using attention or input perturbation or gradient-based measures;~\cite{ghaeini2018interpreting,feng2018pathologies,bach2015pixel}; c) interpretations using surrogate linear models~\cite{ribeiro2016model} -- these methods can provide local interpretations based on input samples or features. However, the presence of inherent randomness makes it difficult to accurately interpret deep neural models among other forms of pathologies~\cite{feng2018pathologies}.
 
 In this paper, we focus on the stability of deep neural models as a function of random-seed based effects. We are especially interested in investigating the hypothesis focusing on model stability: do neural network based models under different random seeds allow for similar interpretations of their decisions? 
 We claim that for a given model which achieves a substantial performance for a task, the factors responsible for any decisions over a sample should be approximately consistent irrespective of the random seed.   
 In Figure~\ref{fig:introeg}, we show an illustration of this question where we visualize the attention distributions of two CNN based binary classification models for sentiment analysis, trained with the same settings and hyper-parameters, but with \textit{different seeds}. We observe that both models obtain the correct prediction with significantly high confidence. However, we note that both the models attend to completely different sets of words. This is problematic, especially when interpreting these models under the influence of such randomness. We observe that on average $40-60\%$ of the most important interpretable units are different across different random seeds for the same model. 
 This phenomenon also leads us to the question on the exact nature of interpretability -- are the interpretations specific to an instantiation of the model or are they general to a class of models?

 We also provide a simple method that can, to a large extent, ameliorate this inherent random behaviour. In Section~\ref{sec:aswa}, we propose an aggressive stochastic weight averaging approach that helps in improving the stability of the models at almost zero performance loss while still making the model robust to random-seed based instability. We also propose an improvement to this model in Section~\ref{ssec:NASWA} which further improves the stability of the neural models. Our proposals significantly improve the robustness of the model, on average by $72\%$ relative to the original model and on Diabetes (MIMIC), a binary classification dataset, by $89\%$ (relative improvement). All code for reproducing and replicating our experiments is released in our repository\footnote{https://github.com/rishj97/ModelStability}.

\section{Measuring Model Stability}
\label{sec:attnstablity}
In this section, we describe methods that we use to measure model stability, specifically --- prediction and interpretation stability. 
\subsection{Prediction Stability} 
We measure prediction stability 
using standard measures of the mean and the standard deviations corresponding to the accuracy of the classification based models on different datasets. We ensure that the models are run with exactly the same configurations and hyper-parameters but with different random seeds. This is a standard procedure that is used in the community to report the performance of the model. 

\subsection{Interpretation Stability}
\label{ssec:attn_stab}
For a given task, we train a set of models only differing with random-seeds. For every given test sample, we obtain interpretations using different instantiations of the models. We define a model to be stable if we obtain similar interpretations regardless of different random-seed based instantiations. We use the following metrics to quantify stability: 

a) \textbf{Relative Entropy quantification ($\mathcal{H}$):} Given two distributions over interpretations, for the same test case, from two different models, it measures the relative entropy between the two probability distributions. Note that, the higher the relative entropy the greater the dissimilarity between the two distributions.
\[
\mathcal{H} = \sum_{i{\in}d}{{{\Pr}_{1}} \cdot \log{\frac{\Pr_1}{\Pr_2}}}
\]
where, $\Pr_1$ and $\Pr_2$ are two attention distributions of the same sample from two different runs of the model and $d$ is the number of tokens in the sample. Given $n$ differently seeded models, for each test instance, we calculate the relative entropy obtained from the corresponding averaged pairwise  interpretation distributions.

b) \textbf{Jaccard Distance ($\mathcal{J}$):} It measures the dissimilarity between two sets. Here higher values of $\mathcal{J}$ indicate larger variances. We consider top-$n$ tokens which have the highest attention for comparison. Note that, Jaccard distance is over sets of word indices and do not take into account the attention probabilities explicitly. Jaccard distance is defined as:
\[
\mathcal{J} = (1 - \frac{A{\cap}B}{A{\cup}B}) * 100 \%
\]
where, $A$ and $B$ are the sets of most relevant items. We specifically decided to use `most' relevant (top-$n$ items) as the tail of the distribution mostly consists of values close to $0$. 

\paragraph{Interpretation methods under study:} In this paper we study interpretation stability using the following three interpretation methods: 

\begin{enumerate} 
\item \textit{Attention based interpretation:} We focus on attention probabilities as the mode of interpretation and consider the model to be stable if different instantiations of the model leads to similar attention distributions. Our major focus in this paper is attention based interpretation. As we use~\newcite{jain2019analysis} as a testbed for our investigation, we focus heavily on attention. Also, as the attention layer has a linear relationship with the prediction, we consider attention to be more indicative of the model stability. 

\item \textit{Gradient-based feature importance:} Given a sample, we use the input gradients of the model corresponding to each of the word representations and compute the magnitude of the change as a local explanation. We refer the reader to~\citet{baehrens2010explain} for a good introduction to gradient-based interpretations. As all of our models are differentiable, we use this as an alternative method for interpretation. We follow the standard procedure as followed in~\citet{feng2018pathologies} and note that we do not follow~\citet{DBLP:journals/corr/abs-1902-10186} and do not disconnect the computational graph at the attention module. We obtain probabilistic gradient scores by normalizing over the absolute values of gradient values.  

\item \textit{LIME based interpretation:} We use locally interpretable model-agnostic interpretations~\cite{ribeiro2016model} that learns a surrogate interpretable model locally around the predictions of the deep neural based model. We obtain LIME based interpretations for every instantiation of the models. We then use Jaccard Distance to measure the divergence.  
\end{enumerate}
We note that, we observe similar patterns across the three interpretation methods and the interpretations consistently differ with random seeds. 
\section{Reducing Model Instability with an Optimization Lens}
We observe that different instantiations of the model can cause the model have different starts on the optimization surface. Further, stochastic sampling might result in different paths. Both of these factors can lead to different local minimas potentially leading to different solutions. With this observation as our background we propose two, closely related, methods to ameliorate divergence: Agressive Stochastic Weight Averaging and Norm-filtered Agressive Stochastic Weight Averaging. We describe these two in the following subsections.  
\subsection{Aggressive Stochastic Weight Averaging (ASWA)} 
\label{sec:aswa}
Stochastic weight averaging (SWA)~\cite{izmailov2018averaging} works by averaging the weights of multiple points in the trajectory of gradient descent based optimizers. The algorithm typically uses modified learning rate schedules. SWA is itself based on the idea of maintaining a running average of weights in stochastic gradient descent based optimization techniques~\cite{ruppert1988stochastic,polyak1992acceleration}. The principle idea in SWA is averaging the weights that are maximally distant helps stabilize the gradient descent based optimizer trajectory and improves generalization. ~\citet{izmailov2018averaging} use the analysis of~\citet{mandt2017stochastic} to illustrate the stability arguments where they show that, under certain convexity assumptions, SGD iterations can be visualized as sampling from a Gaussian distribution centred at the \emph{minima} of the loss function. Samples from high-dimensional Gaussians are expected to be concentrated \emph{on the surface of the ellipse} and not close to the \emph{mean}. Averaging iterations is shown to stabilize the trajectory and further improve the width of the solutions to be closer to the \emph{mean}.

In this paper, we focus on the stability of deep neural models as a function of random-seeds. Our proposal is based on SWA, but we extend it to the extremes and call it \emph{Aggressive Stochastic Weight Averaging}. We assume that, for small batch size, the loss surface is locally convex. We further relax the conditions for the optimizer and assume that the optimizer is based on some version of gradient descent --- this means that our modification is valid even for other pseudo-first-order optimization algorithms including Adam~\cite{kingma2014adam} and Adagrad~\cite{duchi2011adaptive}. 

We note that, \citet{izmailov2018averaging} suggest using SWA usually after 'pre-'training the model (at least until $75\%$ convergence)  and followed by sampling weights at different steps either using large constant or cyclical learning rates. While, SWA is well defined for convex losses~\cite{polyak1992acceleration}, \citet{izmailov2018averaging} connect SWA to non-convex losses by suggesting that the loss surface is~\emph{approximately} convex after convergence.
In our setup, we investigate the utility of averaging weights over every iteration (an iteration consists of one batch of the gradient descent).
Algorithm~\ref{alg:aswa} shows the implementation pseudo-code for SWA. We note that, unlike~\citet{izmailov2018averaging}, we average our weights at \textit{each batch} update and assign the ASWA parameters to the model at the end of each epoch. That is, we replace the model's weights for the next epoch with the averaged weights.

\begin{algorithm}[h]
\begin{algorithmic}[1]
    \Require\\
        $e=$ Epoch number\\
        $m=$ Total epochs\\
        $i=$ Iteration number\\
        $n=$ Total iterations\\
        $\alpha=$ Learning rate\\
        $\mathcal{O}=$ Stochastic Gradient optimizer function
\end{algorithmic}
 $e \gets 0$\;
 \While{$e < m$}{
    {$i \gets 1$}
    
    \While{$i \leq n$}{
      $W_{swa} \gets W_{swa} + \frac{(W - W_{swa})}{(e*n + i + 1)}$\;     
      $W \gets W - \mathcal{O}(\alpha, W)$\;
      $i \gets i + 1$
    }
    $W \gets W_{swa}$\;
    $e \gets e + 1$
 }
 \caption{Aggressive SWA algorithm}
 \label{alg:aswa}
\end{algorithm}

\begin{figure}[h]
 \centering
 \begin{subfigure}{0.88\linewidth}
 \includegraphics[width=\linewidth,keepaspectratio]{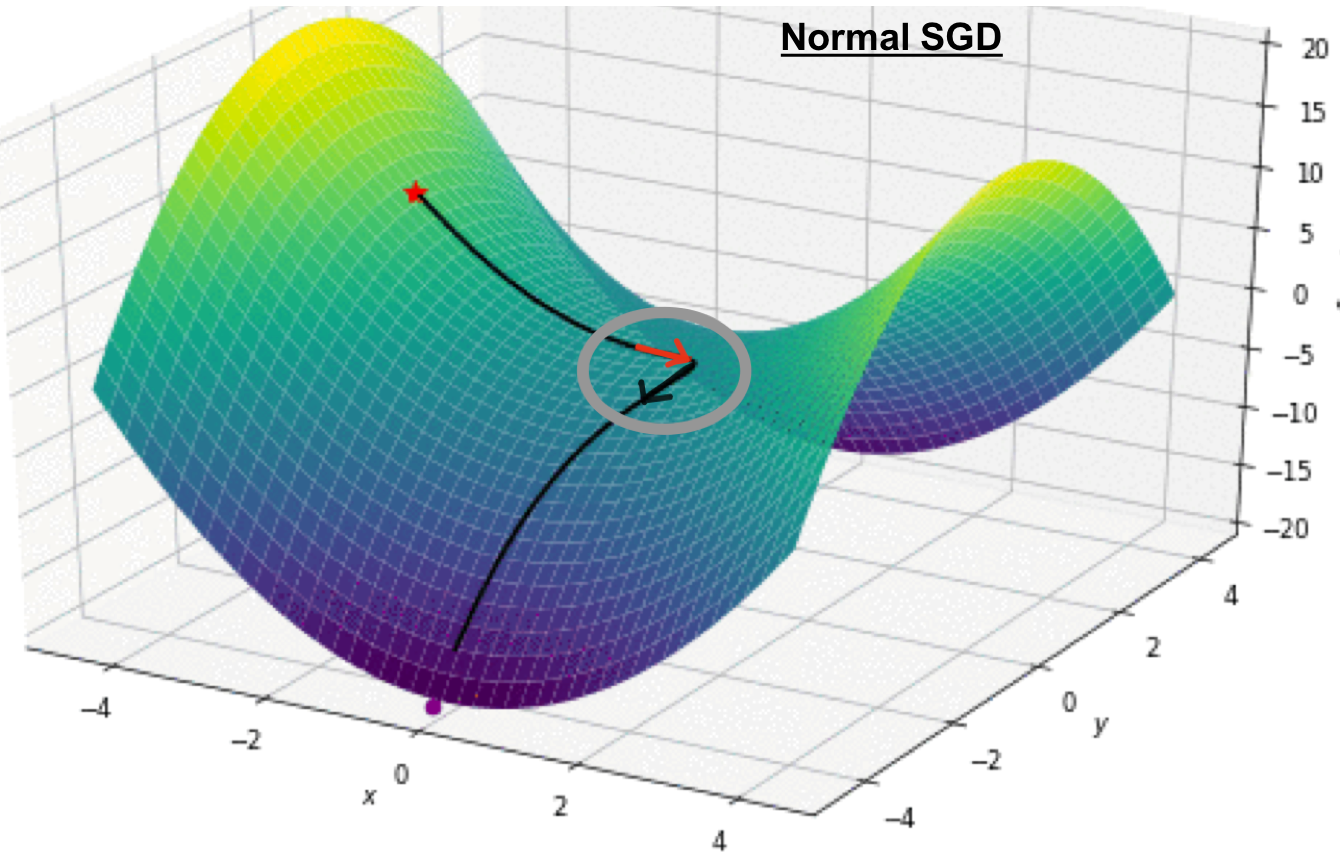}
 \caption{Trajectory for Stochastic Gradient Descent}
 \label{fig:prd_cnn1}
 \end{subfigure}\par\medskip
  \begin{subfigure}{0.88\linewidth}
 \includegraphics[width=\linewidth,keepaspectratio]{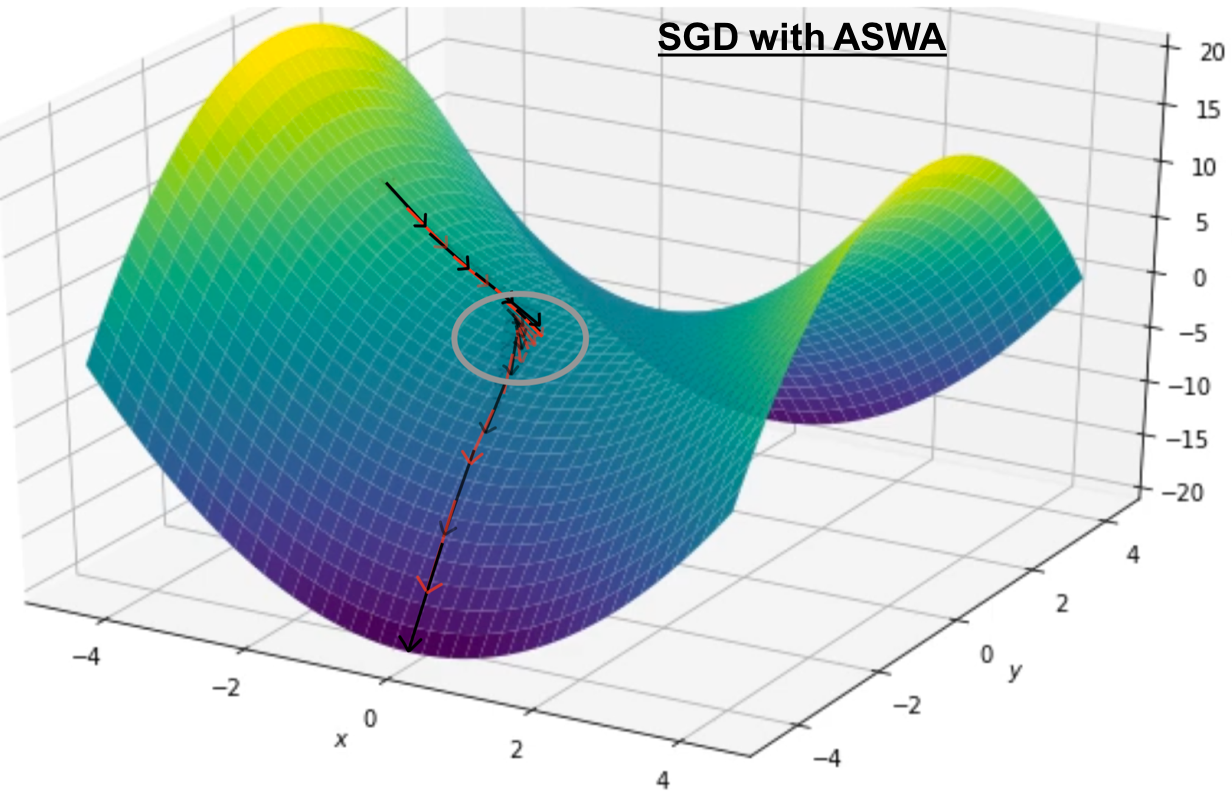}
 \caption{Trajectory for Stochastic Gradient Descent with ASWA}
 \label{fig:prd_cnn2}
 \end{subfigure}
\caption{Trajectory for gradient descent algorithms with red and black arrows on (b) indicating movements from consecutive epochs with restarts. Conservative behaviour of ASWA algorithm helps avoid the saddle point without ever reaching it.}
\label{fig:stable-preds}
\end{figure}

In Figure~\ref{fig:stable-preds}, we show an SGD  optimizer (with momentum) and the same optimizer \emph{with SWA} over a $3$-dimensional loss surface with a saddle point. We observe that the original SGD reaches the desired minima, however, it almost reaches the saddle point and does a course correction and reaches minima. On the other hand, we observe that SGD with ASWA is very conservative, it repeatedly restarts and reaches the minima without reaching the saddle point. We empirically observe that this is a desired property for the stability of models over runs of the same model that differ only over random instantiations. The grey circles in Figure \ref{fig:stable-preds} highlight this conservative behaviour of SGD with ASWA optimizer, especially when compared to the standard SGD. 
Further, ~\citet{polyak1992acceleration} show that for convex losses, averaging SGD proposals achieves the highest possible rate of convergence for a variety of first-order SGD based algorithms.

\subsection{Norm-filtered Aggressive Stochastic Weight Averaging (NASWA)} 
\label{ssec:NASWA}

\begin{algorithm}[h]
\begin{algorithmic}[1]
    \Require\\
        $e=$ Epoch number\\
        $m=$ Total epochs\\
        $i=$ Iteration number\\
        $n=$ Total iterations\\
        $\alpha=$ Learning rate\\
        $\mathcal{O}=$ Stochastic Gradient optimizer function\\
        $N_s=$ List of previous iterations' norm differences
        
\end{algorithmic}
 $e \gets 0$\;
 \While{$e < m$}{
    {$i \gets 1$}
    
    \While{$i \leq n$}{
      $N_{cur} \gets \|W-W_{swa}\|_1$\;
      $N_{mean} \gets \frac{\sum_{i=1}^{|N_s|} N_s[i]}{|N_s|}$\;
      \eIf{$N_{cur} > N_{mean}$}{
      $W_{swa} \gets W_{swa} + \frac{(W - W_{swa})}{(e*n + i + 1)}$\;
      $N_{s} \gets [N_{cur}]$\;
      }{
      $N_{s} \gets N_{s} + [N_{cur}]$\;
      }
      
      $W \gets W - \mathcal{O}(\alpha, W)$\;
      $i \gets i + 1$
    }
    $W \gets W_{swa}$\;
    $e \gets e + 1$
  
 }
\caption{Norm-filtered Aggressive SWA algorithm}
\label{alg:naswa}
\end{algorithm}
We observe that the ASWA algorithm is especially beneficial when the norm difference of the parameters of the model are high. We hypothesise that in general, the norm difference indicates the divergence between optimizers' steps and we observe that the larger the norm difference, the greater the change in the trajectory. Therefore, we propose to maintain a list that stores the norm differences of the previous iterations. If the norm difference of the current iteration is greater than the average of the list, we update the ASWA weights and reinitialize the list with the current norm difference. When the norm difference, however, is less than the average of the list, we just append the current norm difference to the list. After the completion of the epoch, we assign the ASWA parameters to the model. This is shown in Algorithm~\ref{alg:naswa}. We call this approach \emph{Norm-filtered Aggressive Stochastic Weight Averaging}. 
\section{Experiments}
\label{sec:experiemnts}
We base our investigation on similar sets of models as~\citet{DBLP:journals/corr/abs-1902-10186}. We also use the code provided by the authors for our empirical investigations for consistency and empirical validation. We describe our models and datasets used for the experiments below.  
\subsection{Models} 
We consider two sets of commonly used neural models for the tasks of binary classification and multi-class natural language inference. We use CNN and bi-directional LSTM based models with attention. We follow~\cite{DBLP:journals/corr/abs-1902-10186} and use similar attention mechanisms using a) additive attention~\cite{bahdanau2014neural}; and b) scaled dot product based attention~\cite{vaswani2017attention}. We jointly optimize all the parameters for the model, unlike~\citet{DBLP:journals/corr/abs-1902-10186} where the encoding layer, attention layer and the output prediction layer are all optimized separately. We experiment with several optimizers including Adam~\cite{kingma2014adam}, SGD and Adagrad~\cite{duchi2011adaptive} but most results below are with Adam. %

For our ASWA and NASWA based experiments, we use a constant learning rate for our optimizer. Other model-specific settings are kept the same as~\citet{DBLP:journals/corr/abs-1902-10186} for consistency.

\begin{table}[h]
  \centering
  \resizebox{0.9\linewidth}{!}{
  \begin{tabular}{crrr}\toprule
  \textbf{Dataset} & \textbf{Avg. Length} & \textbf{Train Size} & \textbf{Test size}\\
  \midrule
  IMDB & 179 & 12500 / 12500 & 2184 / 2172\\
  Diabetes(MIMIC) & 1858 & 6381 / 1353 & 1295 / 319\\
  SST & 19 & 3034 / 3321 & 652/653 \\
  Anemia(MIMIC) & 2188 & 1847 / 3251 & 460 / 802\\
  AgNews & 36 & 30000 / 30000 & 1900 / 1900\\
  ADR Tweets & 20 & 14446 / 1939 & 3636 / 487\\
  SNLI & 14 & 182764 / 183187 / 183416 &  3219 / 3237 / 3368\\
  \bottomrule
\end{tabular}}
  \caption{Dataset characteristics. Train size and test size show the cardinality for each class. SNLI is a three-class dataset while the rest are binary classification}
\label{tbl:dataset}
\end{table}
\subsection{Datasets}
The datasets used in our experiments are listed in Table~\ref{tbl:dataset} with summary statistics. We further pre-process and tokenize the datasets using the standard procedure and follow~\citet{DBLP:journals/corr/abs-1902-10186}. We note that IMDB~\cite{maas2011learning}, Diabetes(MIMIC)~\cite{johnson2016mimic}, Anemia(MIMIC)~\cite{johnson2016mimic}, AgNews~\cite{zhang2015character}, ADR Tweets~\cite{nikfarjam2015pharmacovigilance} and SST~\cite{socher2013recursive} are datasets for the binary classification setup. SNLI~\cite{bowman2015large} is a dataset for the multiclass classification setup. All of the datasets are in English, however we expect the behavior to persist regardless of the language.%

\subsection{Settings and Hyperparameters}
We use a $300$-dimenstional embedding layer which is initialized with FastText~\cite{joulin2016fasttext} based free-trained embeddings for both CNN and the bi-directional LSTM based models. We use a $128$-dimensional hidden layer for the bi-directional LSTM and a $32$-dimensional filter with kernels of size $\{1,3,5,7\}$ for CNN. For others, we maintain the model settings to resemble the models in~\citet{DBLP:journals/corr/abs-1902-10186}.
We train all of our models for 20 Epochs with a constant batch size of 32. We use early stopping based on the validation set using task-specific metrics (Binary Classification: using \texttt{roc-auc}, Multiclass and question answering based dataset: using \texttt{accuracy}).
\begin{table}[h]
  \centering
  \resizebox{0.9\linewidth}{!}{
  \begin{tabular}{crrr}\toprule
  \textbf{Dataset} & \textbf{CNN(\%)} & \textbf{CNN+ASWA(\%)} & \textbf{CNN+NASWA(\%)}\\
  \midrule
  IMDB & 89.8 ($\pm$0.79) & 90.2 ($\pm$0.25) & 90.1 ($\pm$0.29) \\
  Diabetes & 87.4 ($\pm$2.26) & 85.9 ($\pm$0.25) & 85.9 ($\pm$0.38) \\
  SST & 82.0 ($\pm$1.01) & 82.5 ($\pm$0.39) & 82.5 ($\pm$0.39) \\
  Anemia & 90.6 ($\pm$0.98) & 91.9 ($\pm$0.20) & 91.9 ($\pm$0.19)\\
  AgNews & 95.5 ($\pm$0.23) & 96.0 ($\pm$0.11) & 96.0 ($\pm$0.07)\\
  Tweet & 84.6 ($\pm$2.65) & 84.4 ($\pm$0.54) & 84.4 ($\pm$0.54)
  \\\bottomrule
  \end{tabular}}
  \caption{Performance statistics obtained from 10 differently seeded CNN based models. Table compares accuracy and its \textbf{standard deviation} for the normally trained CNN model against the ASWA and NASWA trained models, whose deviation drops significantly, thus, indicating increased robustness.}
  \label{tbl:acc_std}
\end{table}

\section{Results}
In this section, we summarize our findings for $10$ runs of the model with $10$ different random seeds but with identical model settings.

\subsection{Model Performance and Stability}

We first report model performance and prediction stability. The results are reported in Table~\ref{tbl:acc_std}.  
\begin{table}[!h]
  \centering
  \resizebox{0.9\linewidth}{!}{
  \begin{tabular}{crrr}\toprule
  \textbf{Dataset} & \textbf{LSTM(\%)} & \textbf{LSTM+ASWA(\%)} & \textbf{LSTM+NASWA(\%)}\\
  \midrule
  IMDB & 89.1 ($\pm$1.34) & 90.2 ($\pm$0.32) & 90.3 ($\pm$0.17) \\
  Diabetes & 87.7 ($\pm$1.44) & 87.7 ($\pm$0.60) & 87.8 ($\pm$0.55) \\
  SST & 81.9 ($\pm$1.11) & 82.0 ($\pm$0.60) & 82.1 ($\pm$0.57) \\
  Anemia & 91.6 ($\pm$0.49) & 91.8 ($\pm$0.34) & 91.9 ($\pm$0.36)\\
  AgNews & 95.5 ($\pm$0.32) & 96.1($\pm$0.17) & 96.1 ($\pm$0.10)\\
  Tweet & 84.7 ($\pm$1.79) & 83.8 ($\pm$0.45) & 83.9 ($\pm$0.45)
  \\\bottomrule
  \end{tabular}}
  \caption{Performance statistics obtained from 10 differently seeded LSTM based models.}
  \label{tbl:acc_std_lstm}
\end{table}

We note that the original CNN based models, on an average, have a standard deviation of $\pm 1.5\%$. Which seems standard, however, we note that ADR Tweets~dataset has a very high standard deviation of $\pm 2.65\%$. We observe that ASWA and NASWA are almost always able to achieve higher performance with a very low standard deviation. This suggests that both ASWA and NASWA are extremely stable when compared to the standard model. They significantly improve the robustness, on an average, by $72\%$ relative to the original model and on Diabetes (MIMIC), a binary classification dataset, by $89\%$ (relative improvement). We observe similar results for the LSTM based models in Table~\ref{tbl:acc_std_lstm}.

\begin{figure}[h]
 \centering
 \begin{subfigure}{0.5\linewidth}
 \includegraphics[width=\linewidth,keepaspectratio]{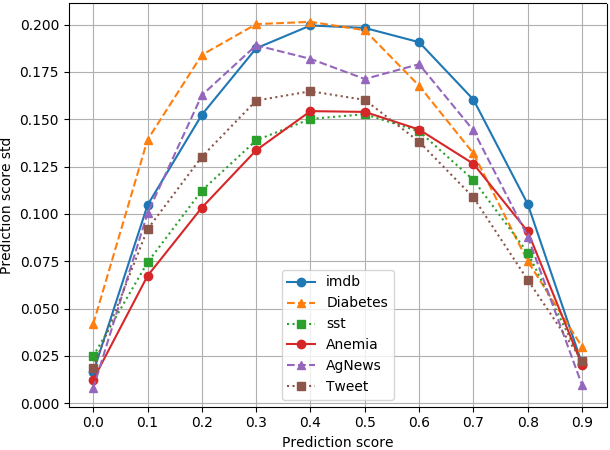}
 \caption{CNN models}
 \label{fig:prd_cnn}
 \end{subfigure}\hfill
  \begin{subfigure}{0.5\linewidth}
 \includegraphics[width=\linewidth,keepaspectratio]{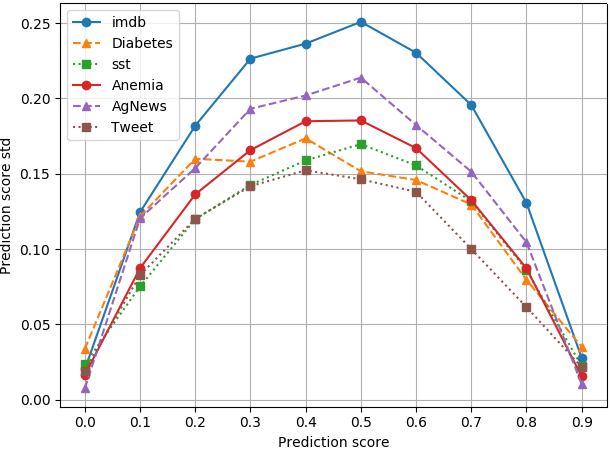}
 \caption{LSTM models}
 \label{fig:prd_lstm}
 \end{subfigure}
\caption{Prediction's standard deviation for CNN and LSTM based models for all binary classification datasets under consideration. Predictions are bucketed in intervals of size 0.1, starting from 0 (containing predictions from 0 to 0.1), until 0.9} 
\end{figure}

We further analyze the prediction score stability by computing the mean standard deviation over the binned confidence intervals of the models in Figure~\ref{fig:prd_cnn}.  We note that on an average, the standard deviations are on the lower side. However, we observe that the mean standard deviation of the bins close to $0.5$ is on the higher side as is expected given the high uncertainty.  On the other hand both, ASWA and NASWA based models are relatively more stable than the standard CNN based model. We observe similar behaviours for the LSTM based models in Figure~\ref{fig:prd_lstm}. 
This suggests that our proposed methods, ASWA and NASWA, are able to obtain relatively better stability without any loss in performance. We also note that both ASWA and NASWA had relatively similar performance over more than $10$ random seeds.

\subsection{Attention Stability}
\begin{figure}[h]
 \begin{subfigure}{0.5\linewidth}
 \includegraphics[width=\linewidth,keepaspectratio]{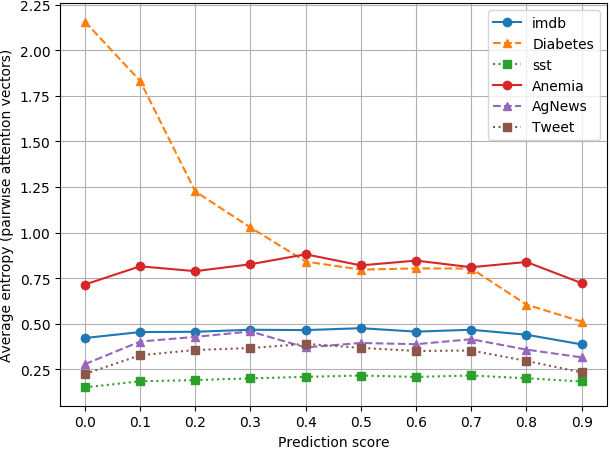}
 \caption{CNN models}
 \label{fig:atn_cnn}
 \end{subfigure}\hfill
  \begin{subfigure}{0.5\linewidth}
 \includegraphics[width=\linewidth,keepaspectratio]{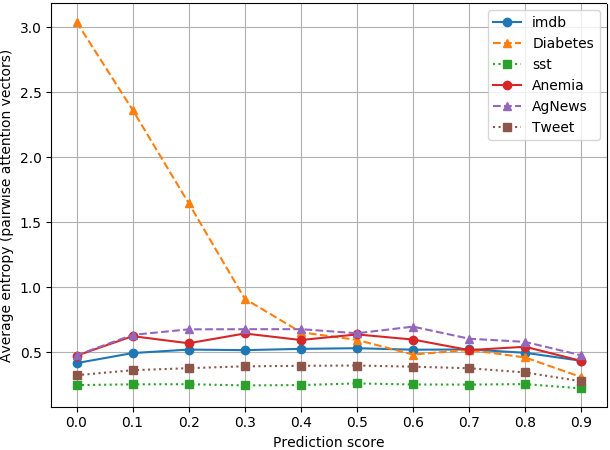}
 \caption{LSTM models}
 \label{fig:atn_lstm}
 \end{subfigure}
 \caption{Average attention entropy against the bucketed predictions for CNN and LSTM based models. Figure highlights the high entropy between attention based distributions from differently seeded models (especially for the Diabetes-MIMIC datatset), indicating towards model instability.}
\end{figure}

We now consider the stability of attention distributions as a function of random seeds. We first plot the results of the experiments for \emph{standard} CNN based binary classification models over uniformly binned prediction scores for positive labels in Figure~\ref{fig:atn_cnn}. We observe that, depending on the datasets, the attention distributions can become extremely unstable (high entropy). We specifically highlight the Diabetes(MIMIC) dataset's entropy distribution. 
We observe similar, but relatively worse results for the LSTM based models in Figure~\ref{fig:atn_lstm}. In general, we would expect the entropy distribution to be close to zero however, this doesn't seem to be the case. This means that using attention distributions to interpret models may not be reliable and can lead to misinterpretations.

\begin{figure}[h]
 \centering
 \begin{subfigure}{0.5\linewidth}
 \includegraphics[width=\linewidth,keepaspectratio]{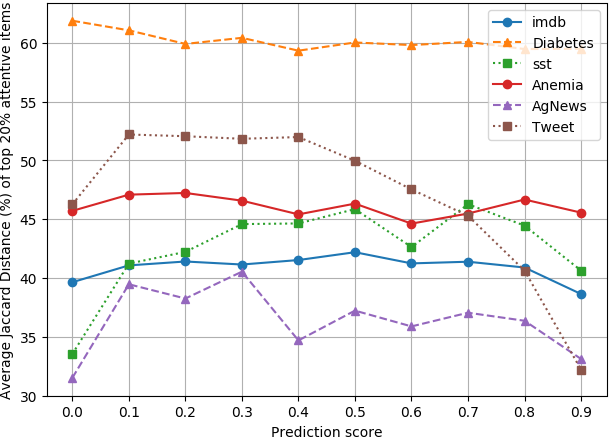}
 \caption{CNN models}
 \label{fig:std_jaccard_cnn}
 \end{subfigure}\hfill
 \begin{subfigure}{0.5\linewidth}
 \includegraphics[width=\linewidth,keepaspectratio]{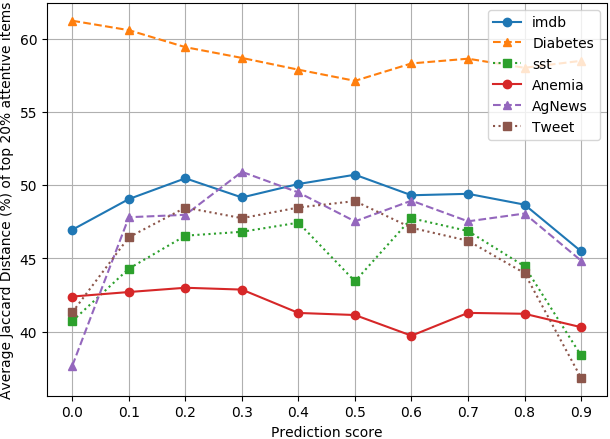}
  \caption{LSTM models}
 \label{fig:std_jaccard_lstm}
 \end{subfigure}
 \caption{Jaccard distance highlighting instability in attention distributions of CNN and LSTM based models.}
\end{figure}

\begin{figure}[h]
 \centering
 \begin{subfigure}{0.5\linewidth}
 \includegraphics[width=\linewidth,keepaspectratio]{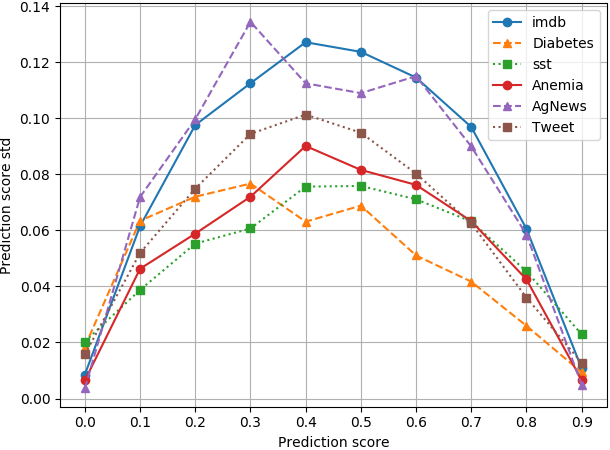}
 \caption{CNN+ASWA}
 \end{subfigure}\hfill
 \begin{subfigure}{0.5\linewidth}
 \includegraphics[width=\linewidth,keepaspectratio]{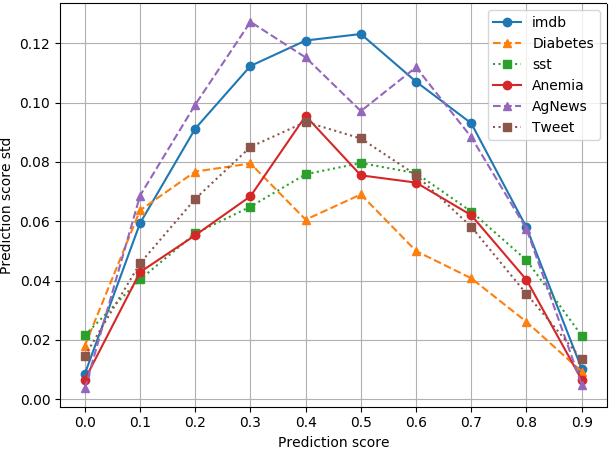}
  \caption{CNN+NASWA}
 \end{subfigure}\par\medskip
 \begin{subfigure}{0.5\linewidth}
 \includegraphics[width=\linewidth,keepaspectratio]{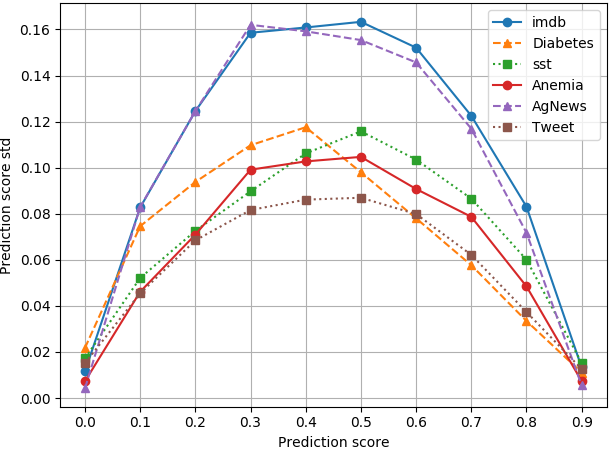}
 \caption{LSTM+ASWA}
 \end{subfigure}\hfill
 \begin{subfigure}{0.5\linewidth}
 \includegraphics[width=\linewidth,keepaspectratio]{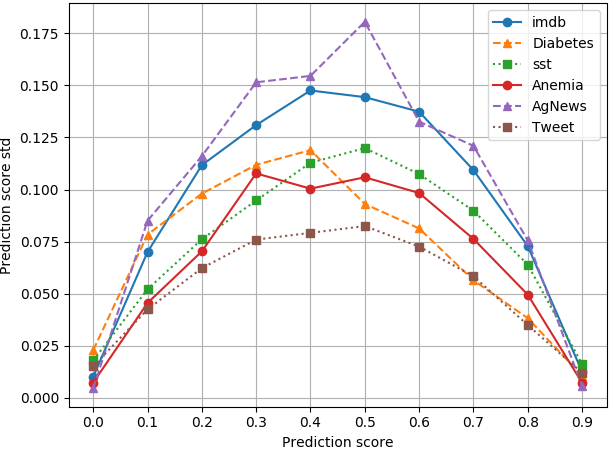}
  \caption{LSTM+NASWA}
 \end{subfigure}
 \caption{Improved prediction stability from ASWA and NASWA for CNN and LSTM based models}
\end{figure}
\begin{figure}[h]
  \begin{subfigure}{0.45\linewidth}
  \centering
  \includegraphics[width=\linewidth,keepaspectratio]{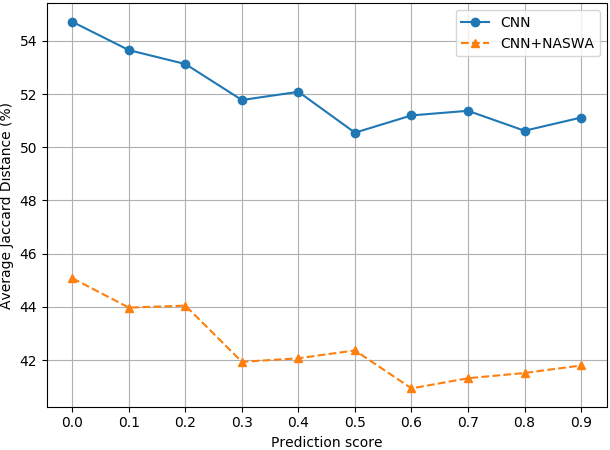}\hfill
  \caption{Diabetes}
  \end{subfigure}\hfill
  \begin{subfigure}{0.45\linewidth}
  \centering
  \includegraphics[width=1.0\linewidth,keepaspectratio]{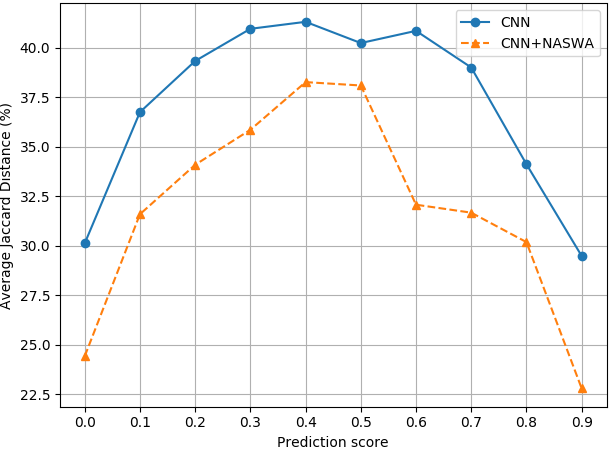}\hfill
  \caption{SST}
  \end{subfigure}
\caption{Gradient based interpretations' stability improvement from NASWA on CNN based models. The Jaccard distance is calculated using the top 20\% attentive items.}
\label{fig:grad_stab}
\end{figure}

\begin{figure*}
  \begin{subfigure}{0.3\linewidth}
  \centering
  \includegraphics[width=\linewidth,keepaspectratio]{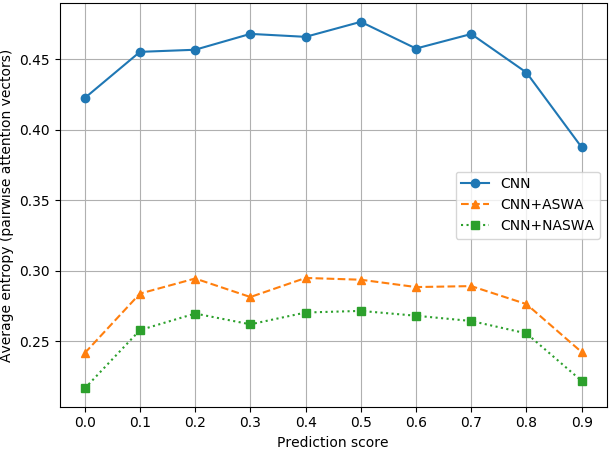}\hfill
  \caption{IMDB}
  \label{sfig:aswa_imdb}
  \end{subfigure}\hfill
  \begin{subfigure}{0.3\linewidth}
  \centering
  \includegraphics[width=1.0\linewidth,keepaspectratio]{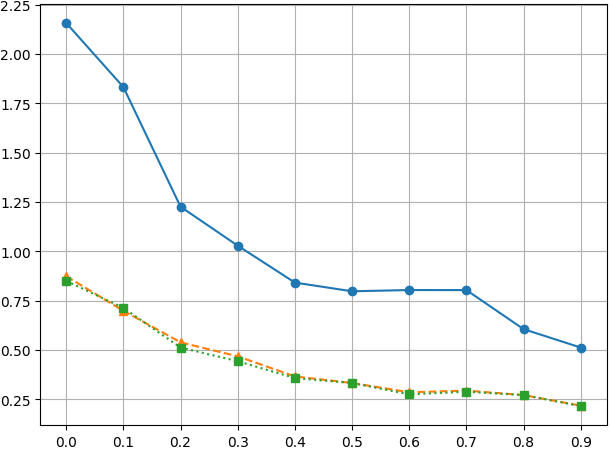}\hfill
  \caption{Diabetes}
  \label{sfig:aswa_diab}
  \end{subfigure}\hfill
  \begin{subfigure}{0.3\linewidth}
  \centering
  \includegraphics[width=1.0\linewidth,keepaspectratio]{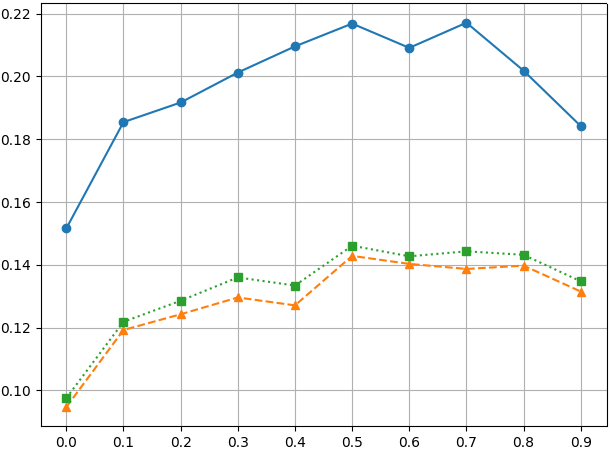}\hfill
  \caption{SST}
  \end{subfigure}\par\medskip
  \begin{subfigure}{0.3\linewidth}
  \centering
  \includegraphics[width=1.0\linewidth,keepaspectratio]{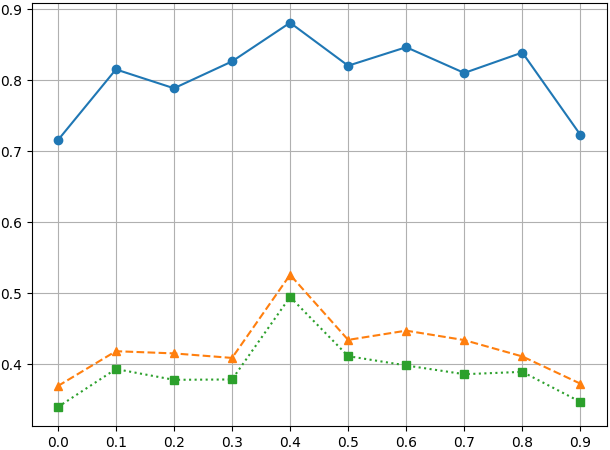}\hfill
  \caption{Anemia}
  \end{subfigure}\hfill
    \begin{subfigure}{0.3\linewidth}
  \centering
  \includegraphics[width=1.0\linewidth,keepaspectratio]{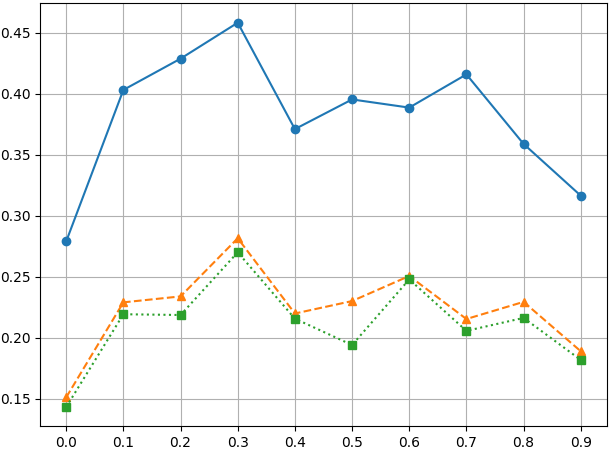}\hfill
  \caption{AgNews}
  \end{subfigure}\hfill
    \begin{subfigure}{0.3\linewidth}
  \centering
  \includegraphics[width=1.0\linewidth,keepaspectratio]{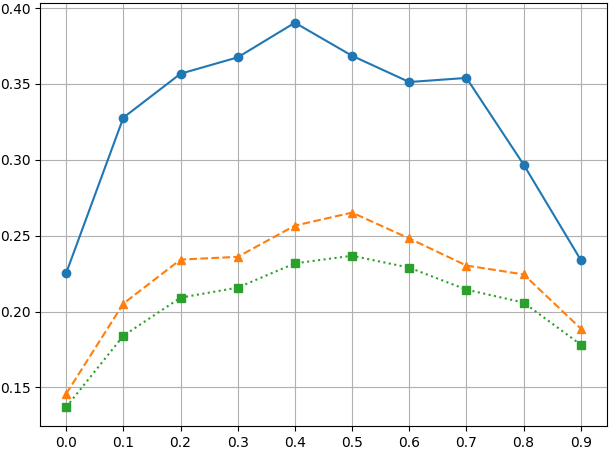}\hfill
  \caption{ADR Tweets}
  \end{subfigure}
\caption{Attention stability improvement from ASWA and NASWA on CNN based models.}
\label{fig:attn_stab_imp}
\end{figure*}
\begin{figure*}[!h]
  \begin{subfigure}{0.3\linewidth}
  \centering
  \includegraphics[width=\linewidth,keepaspectratio]{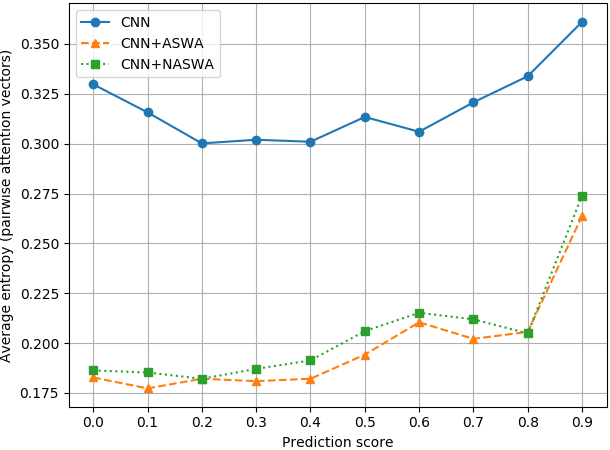}\hfill
  \caption{Label 0 prediction vs entropy}
  \end{subfigure}\hfill
  \begin{subfigure}{0.3\linewidth}
  \centering
  \includegraphics[width=1.0\linewidth,keepaspectratio]{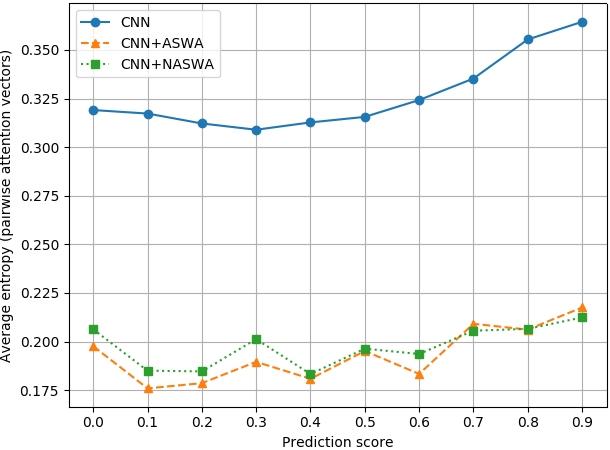}\hfill
  \caption{Label 1 prediction vs entropy}
  \end{subfigure}\hfill
  \begin{subfigure}{0.3\linewidth}
  \centering
  \includegraphics[width=1.0\linewidth,keepaspectratio]{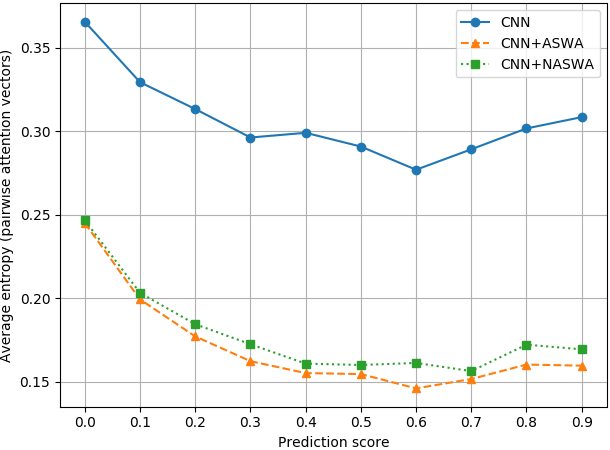}\hfill
  \caption{Label 2 prediction vs entropy}
  \end{subfigure}
\caption{Attention stability improvement from ASWA and NASWA on CNN based model for the SNLI dataset.}
\label{fig:snli_cnn_entropy}
\end{figure*}

We use the top $20\%$ of the most important items (indices) in the attention distribution for each dataset over $10$ runs and plot the Jaccard distances for CNN and LSTM based models in Figure~\ref{fig:std_jaccard_cnn} and Figure~\ref{fig:std_jaccard_lstm}. We again notice a similar trend of unstable attention distributions over both CNN and LSTM based attention distribution. 

In the following sections for space constraints, we focus on CNN based models with additive attention. Our results on LSTM based models are provided in the attached supplementary material. We note that the observations for LSTM models are, in most cases, similar to the behaviour of the CNN based models. Scaled dot-product based models are also provided in the supplementary material and we notice a similar trend as the additive attention.

We now focus on the effect of ASWA and NASWA on binary and multi-class CNN based neural models separately.

\paragraph{Binary Classification}

In Figure~\ref{fig:attn_stab_imp}, we plot the results of the models with ASWA and NASWA. We observe that both these algorithms significantly improve the model stability and decrease the entropy between attention distributions. For example, in Figure~\ref{sfig:aswa_diab}, both ASWA and NASWA decrease the average entropy by about $60\%$. We further notice that NASWA is slightly better performing in most of the runs. This empirically validates the hypothesis that averaging the weights from divergent weights (when the norm difference is higher than the average norm difference) helps in stabilizing the model's parameters, resulting in a more robust model.
\paragraph{Multi-class Classification}
In Figure \ref{fig:snli_cnn_entropy}, we plot the entropy between the attentions distributions of the models for the SNLI dataset (CNN based model), separately for \emph{each label} (\emph{neutral}, \emph{contradiction}, and \emph{entailment}). We notice, similar observations as the binary classification models, the ASWA and NASWA algorithms are able to significantly improve the entropy of the attention distributions and increases the robustness of the model with random seeds. 
\subsection{Gradient-based Interpretations}
We now look at an alternative method of interpreting deep neural models and look into the consistency of the gradient-based interpretations to further analyze the model's instability. For this setup, we focus on binary classifier and plot the results on the SST and the Diabetes dataset in particular since they cover the low and the high end of the entropy spectrum (respectively).
We notice similar trends of instability in the gradient-based interpretations from model inputs as we did for the attention distributions. Figure \ref{fig:grad_stab} shows that the entropy between the gradient-based interpretations from differently seeded models closely follows the same trend as the attention distributions. 
This result further strengthens our claim on the importance of model stability and shows that over different runs of the same model with different seeds, we may get different interpretations using gradient-based feature importance. Moreover, Figure \ref{fig:grad_stab} shows the impact of ASWA towards making the gradient-based interpretations more consistent, thus, significantly increasing the stability.

\subsection{LIME based Interpretations}
\begin{figure}
  \begin{subfigure}{0.5\linewidth}
  \centering
  \includegraphics[width=\linewidth,keepaspectratio]{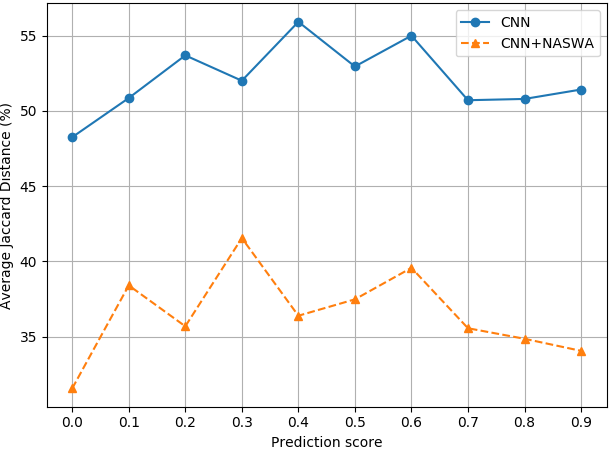}\hfill
  \caption{AgNews}
  \end{subfigure}\hfill
  \begin{subfigure}{0.5\linewidth}
  \centering
  \includegraphics[width=1.0\linewidth,keepaspectratio]{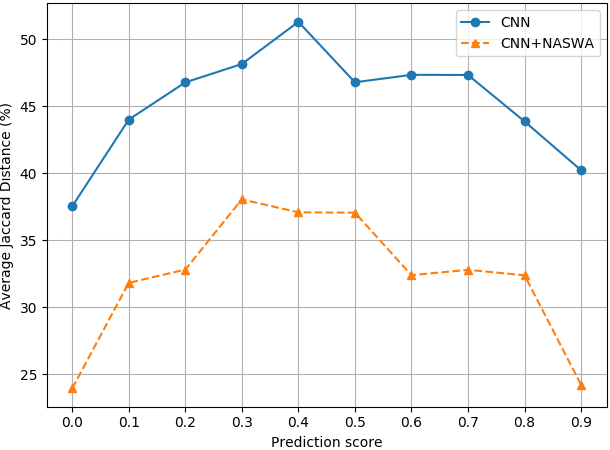}\hfill
  \caption{SST}
  \end{subfigure}
\caption{LIME based interpretations' stability improvement from NASWA on CNN based models. The Jaccard distance is calculated using the top 20\% attentive items.}
\label{fig:lime_stab}
\end{figure}

We further evaluated the surrogate model based interpretability using LIME~\cite{ribeiro2016model}.  LIME obtains a locally linear approximation of the model's behaviour for a given sample by perturbing it and learning a sparse linear model around it. We focus on AgNews and SST based datasets and obtain interpretability estimates using LIME. Once again, we notice a similar pattern of instability as the other two interpretability methods. In Figure~\ref{fig:lime_stab} we present our results from the LIME based interpretations with Jaccard distance as the measure. Note that we measure the Jaccard distance over the top $20\%$ most influential items. We observe once again that NASWA helps in reducing the instability and results in more consistent interpretations. 

In all our experiments, we find that a significant proportion of interpretations are dependent on the instantiation of the model. We also note that we perform experiments over $100$ random seeds for greater statistical power and see similar patterns\footnote{These results are provided in the appendix.}.  %
\section{Discussion}
Recent advances in adversarial machine learning~\cite{neelakantan2015adding,zahavy2016ensemble} have investigated robustness to random initialization based perturbations, however, to our knowledge, no previous study investigates the effect of random-seeds and its connection on model interpretation. Our study analyzed the inherent lack of robustness in deep neural models for NLP. Recent studies cast doubt on the consistency and correlations of several types of interpretations~\cite{doshi2017towards,DBLP:journals/corr/abs-1902-10186,feng2018pathologies}. We hypothesise that some of these issues are due to the inherent instability of the deep neural models to random-seed base perturbations. %
Our analysis (in Section~\ref{sec:experiemnts}) leads to the hypothesis that models with different instantiations may use completely different optimization paths. The issue of variance in all black-box interpretation methods over different seeds will continue to persist until the models are fully robust to random-seed based perturbations. Our work however, doesn't provide insights into instabilities of different layers of the models. We hypothesise that it might further uncover the reasons for the relatively lower correlation between different black-box interpretation methods as these are effectively based off on different layers and granularity. 

There has been some work on using noisy gradients~\cite{neelakantan2015adding} and learning from adversarial and counter-factual examples~\cite{feng2018pathologies} to increase the robustness of deep learning models.~\citet{feng2018pathologies} show that neural models may use redundant features for prediction and also show that most of the black-box interpretation methods may not be able to capture these second-order effects. Our proposals show that aggressively averaging weights leads to better optimization and the resultant models are more robust to random-seed based perturbation. However, our research is limited to increasing consistency in neural models. Our approach further uses first order based signals to boost stability. We posit that second-order based signals can further enhance consistency and increase the robustness.%

\section{Conclusions}
In this paper, we study the inherent instability of deep neural models in NLP as a function of random seed. We analyze model performance and robustness of the model in the form of attention based interpretations, gradient-based feature importance and LIME based interpretations across multiple runs of the models with different random seeds. Our analysis strongly highlights the problems with stability of models and its effects on black-box interpretation methods leading to different interpretations for different random seeds. 
We also propose a solution that makes use of weight averaging based optimization technique and further extend it with norm-filtering. We show that our proposed methods largely stabilize the model to random-seed based perturbations and, on average, significantly reduce the standard deviations of the model performance by $72\%$. We further show that our methods significantly reduce the entropy in the attention distribution, the gradient-based feature importance measures and LIME based interpretations across runs.

  \section*{Acknowledgments}
We thank Panos Parpas and Emtiyaz Khan for their feedback on an earlier draft of this paper. We thank the anonymous reviewers for their thorough reviews and constructive comments. Pranava Madhyastha kindly acknowledges the support of Amazon AWS Cloud Credits for Research Award, hardware grant from NVIDIA, Anne O'Neill and the Imperial Corporate Partnership Programme. 
\bibliography{refs}
\bibliographystyle{acl_natbib}
\clearpage
\appendix

\begin{figure}[!h]
 \centering
 \resizebox{.9\linewidth}{!}{
 \includegraphics[width=\linewidth,keepaspectratio]{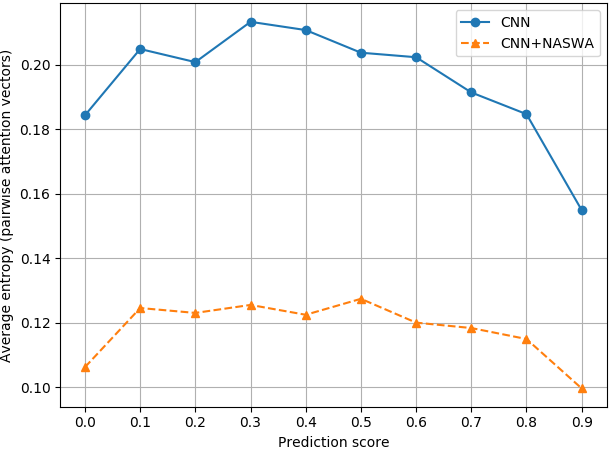}
 }
 \caption{Entropy improvement for tanh Attention based CNN model for the SST dataset using 100 different seeds.}
 \label{fig:sst-100}
\end{figure}

\begin{figure}[!h]
 \centering
 \resizebox{.9\linewidth}{!}{
 \includegraphics[width=\linewidth,keepaspectratio]{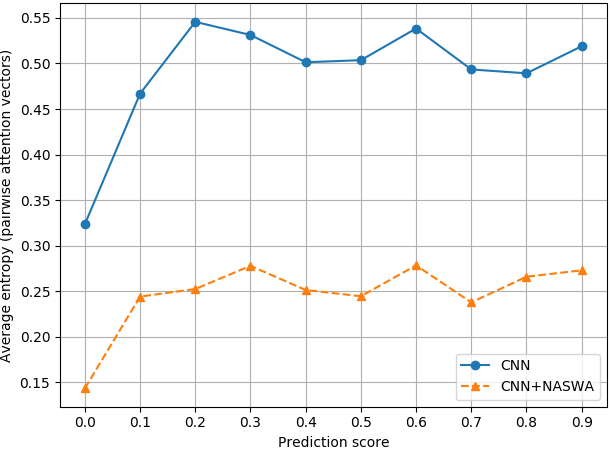}
 }
 \caption{Entropy improvement for tanh Attention based CNN model for the AgNews dataset using 100 different seeds.}
 \label{fig:agnews-100}
\end{figure}

\begin{figure}[!h]
 \centering
 \resizebox{.9\linewidth}{!}{
 \includegraphics[width=\linewidth,keepaspectratio]{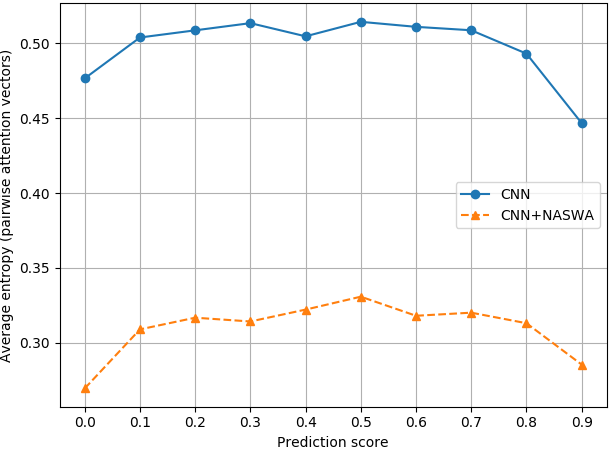}
 }
 \caption{Entropy improvement for tanh Attention based CNN model for the IMDB dataset using 100 different seeds.}
 \label{fig:imdb-100}
\end{figure}
\begin{figure*}[!h]
  \begin{subfigure}{0.3\linewidth}
  \centering
  \includegraphics[width=\linewidth,keepaspectratio]{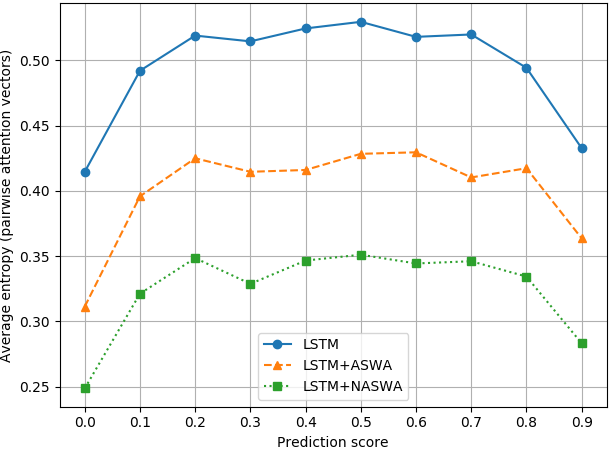}\hfill
  \caption{IMDB}
  \end{subfigure}\hfill
  \begin{subfigure}{0.3\linewidth}
  \centering
  \includegraphics[width=1.0\linewidth,keepaspectratio]{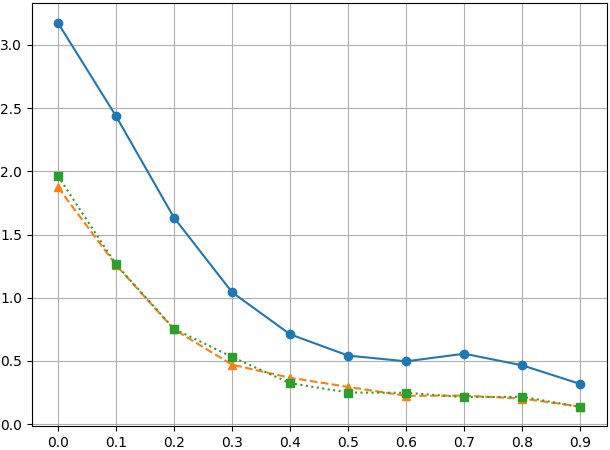}\hfill
  \caption{Diabetes}
  \end{subfigure}\hfill
  \begin{subfigure}{0.3\linewidth}
  \centering
  \includegraphics[width=1.0\linewidth,keepaspectratio]{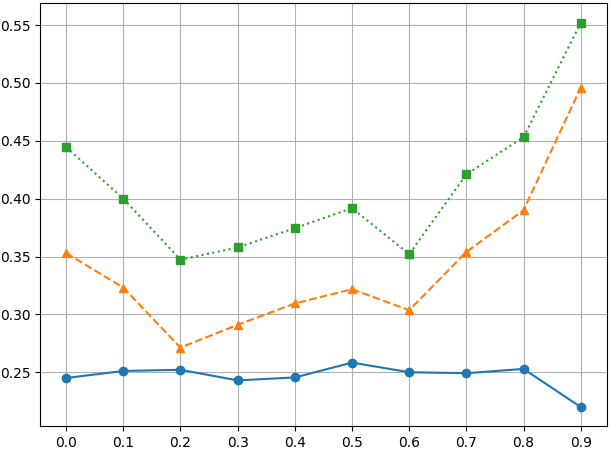}\hfill
  \caption{SST}
  \end{subfigure}\par\medskip
  \begin{subfigure}{0.3\linewidth}
  \centering
  \includegraphics[width=1.0\linewidth,keepaspectratio]{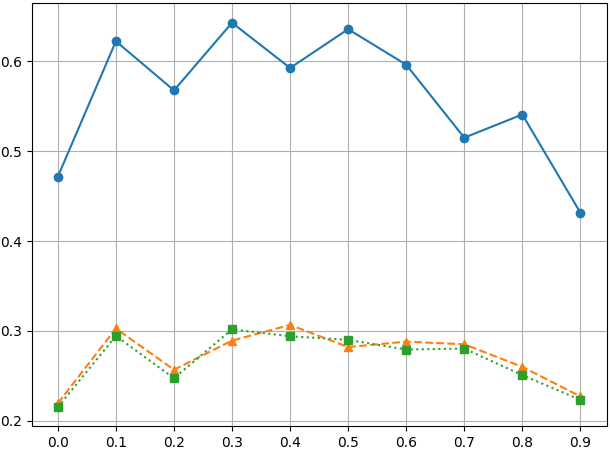}\hfill
  \caption{Anemia}
  \end{subfigure}\hfill
    \begin{subfigure}{0.3\linewidth}
  \centering
  \includegraphics[width=1.0\linewidth,keepaspectratio]{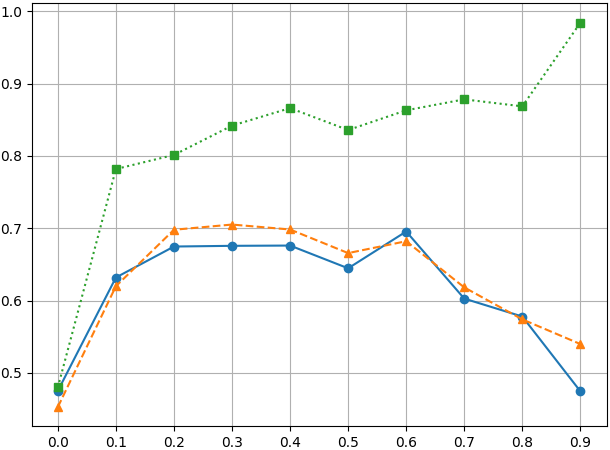}\hfill
  \caption{AgNews}
  \end{subfigure}\hfill
    \begin{subfigure}{0.3\linewidth}
  \centering
  \includegraphics[width=1.0\linewidth,keepaspectratio]{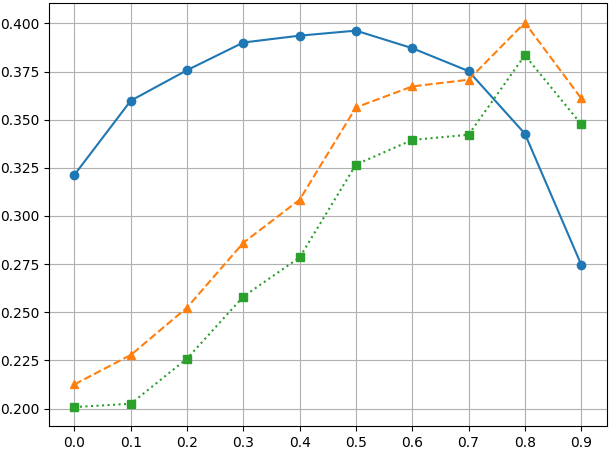}\hfill
  \caption{ADR Tweets}
  \end{subfigure}
\caption{Attention stability improvement from ASWA and NASWA on LSTM based models.}
\label{fig:lstm-atn-entropy}
\end{figure*}

\section{Jaccard Distance Experiments}
In Figure~\ref{fig:std_jaccard_cnn} and Figure~\ref{fig:std_jaccard_lstm} we plot the Jaccard distance plots with CNN models and LSTM models and note that ASWA consistently improves the stability of the interpretations. 
\begin{figure}[!h]
 \centering
 \resizebox{.9\linewidth}{!}{
 \includegraphics[width=\linewidth,keepaspectratio]{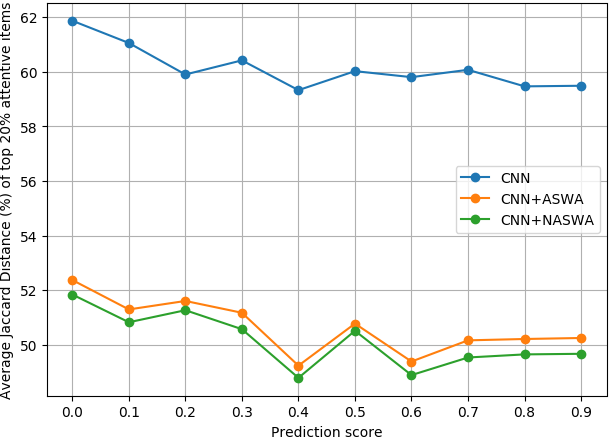}
 }
 \caption{Jaccard Distance improvement for Diabetes.}
\end{figure}

\begin{figure}[!h]
 \centering
 \resizebox{.9\linewidth}{!}{
 \includegraphics[width=\linewidth,keepaspectratio]{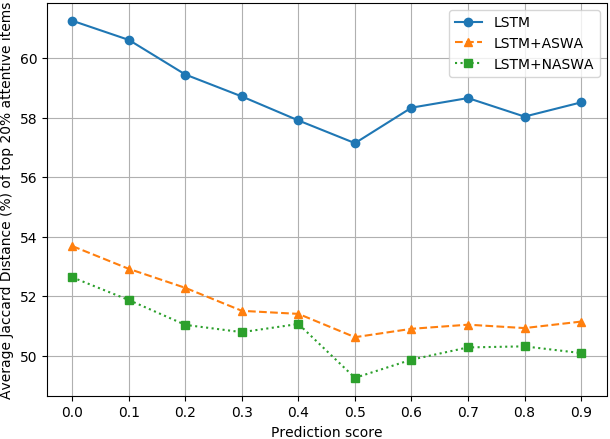}
 }
 \caption{Jaccard Distance improvement for Diabetes.}
\end{figure}

\begin{figure}[!h]
 \centering
 \resizebox{.9\linewidth}{!}{
 \includegraphics[width=\linewidth,keepaspectratio]{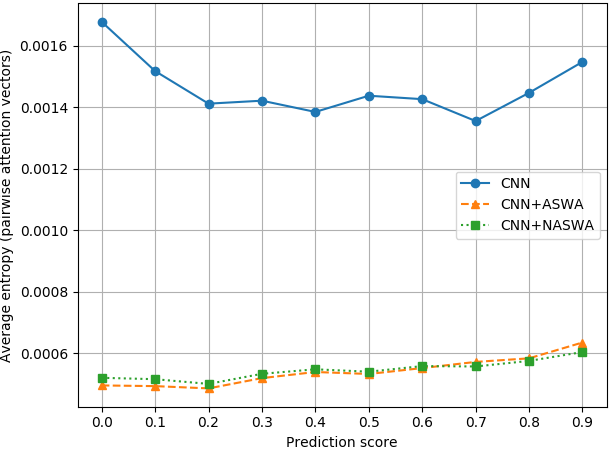}
 }
 \caption{Entropy improvement for dot Attention based CNN model for SST dataset.}
\end{figure}

\begin{figure}[!h]
 \centering
 \resizebox{.9\linewidth}{!}{
 \includegraphics[width=\linewidth,keepaspectratio]{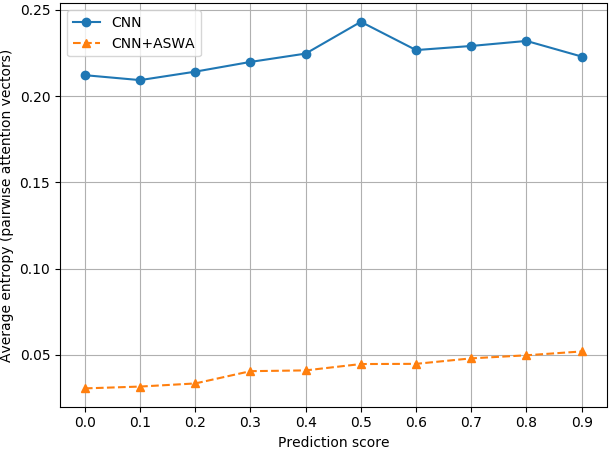}
 }
 \caption{Entropy improvement for dot Attention based CNN model for Diabetes dataset.}
\end{figure}

\section{Results with 100 seeds}
We perform attention stability based experiments as mentioned in the paper, but now with 100 seeds, i.e, 100 models initialized with different seeds instead of 10. Figures \ref{fig:sst-100}, \ref{fig:imdb-100}, and \ref{fig:agnews-100} show the entropy of attention based interpretations for different datasets. Experimenting with 100 seeds helps strengthen our claims about the instability of the model and the effectiveness of our proposed algorithms like ASWA and NASWA.

\section{Hyperparameter Settings}
For training purposes, we use the same model settings for the models as mentioned in the paper \cite{DBLP:journals/corr/abs-1902-10186} (or the Github implementation\footnote{\url{https://github.com/successar/AttentionExplanation}}), our port of the code is made available at: \url{https://github.com/rishj97/ModelStability}. Additional hyper-parameters for replication studies are: 
\begin{itemize}
    \item Number of epochs: 20
    \item Optimizer: Adam
    \item Learning rate: 0.001
\end{itemize}
The exact seeds used for running the experiments can be found in our code repository.

\section{Binary Classification with LSTM based Models}
For LSTM based models, we notice (in Figure \ref{fig:lstm-atn-entropy}) similar trends as to the CNN models in terms of the instability of the attention based interpretations.

\end{document}